\documentclass[10pt,twocolumn,letterpaper]{article}

\usepackage{cvpr}
\usepackage{times}
\usepackage{epsfig}
\usepackage{graphicx}
\usepackage{amsmath}
\usepackage{amssymb}
\usepackage{algorithm}
\usepackage[noend]{algpseudocode}
\graphicspath{ {img/} }

\usepackage[pagebackref=true,breaklinks=true,letterpaper=true,colorlinks,bookmarks=false]{hyperref}

\cvprfinalcopy 

\def\cvprPaperID{****} 

\ifcvprfinal\fi
\begin{document}

\title{3D Traffic Simulation for Autonomous Vehicles in Unity and Python}

\author{Zhijing Jin\\
The University of Hong Kong\\
Pok Fu Lam, Hong Kong\\
{\tt\small zhijing.jin@connect.hku.hk}
\and
Tristan Swedish\\
Massachusetts Institute of Technology\\
Cambridge, MA, USA 02139\\
{\tt\small tswedish@mit.edu}
\and
Ramesh Raskar\\
Massachusetts Institute of Technology\\
Cambridge, MA, USA 02139\\
{\tt\small raskar@mit.edu}
}

\maketitle

\begin{abstract}
   Over the recent years, there has been an explosion of studies on autonomous vehicles. Many collected large amount of data from human drivers. However, compared to the tedious data collection approach, building a virtual simulation of traffic makes the autonomous vehicle research more flexible, time-saving, and scalable. Our work features a 3D simulation that takes in real time position information parsed from street cameras. The simulation can easily switch between a global bird view of the traffic and a local perspective of a car. It can also filter out certain objects in its customized camera, creating various channels for objects of different categories. This provides alternative supervised or unsupervised ways to train deep neural networks. Another advantage of the 3D simulation is its conformation to physical laws. Its naturalness to accelerate and collide prepares the system for potential deep reinforcement learning needs.\footnote{Code is available at https://github.com/zhijing-jin/3D\_Traffic\_in\_Unity}
\end{abstract}

\section{Introduction}

For years people have been trying to tackle the problem of autonomous vehicles. The use of autonomous vehicles will have a huge impact on traffics. Research has found that over 90\% car accidents are identified as human errors, compared to the 2\% vehicle failures \cite{singh2015critical}. The motivation of research on self-driving vehicles lies in both safety concerns and improvement of traffic flow and emissions. Mainstream companies including Google, Toyota, Nissan and Audi have put much effort into the development of autonomous vehicle prototypes \cite{bansal2016assessing}.

Recent breakthrough of autonomous vehicles is based on the use of deep neural networks such as Convolutional neural networks (CNNs), Recurrent Neural Networks (RNNs) and Deep Reinforcement Learning. The decision making process of an autonomous vehicle is usually formulated as a computer vision problem, which outputs a manipulation action of vehicle based on the previous few frames collected by the vehicle camera. For example, NVIDIA has deployed an end-to-end approach to map raw pixels from a car camera to steering commands \cite{bojarski2016end}. 

However, a large problem for training deep neural networks is the lack of data. As deep neural networks have millions of parameters, a large number of good data is decisive for successful models. Many companies collect data from cameras on vehicles that are driven long distance on highways and streets. Concerns are that these data are expensive to collect and also lack edge cases. Here, we propose a new way of collecting traffic data from street cameras. This source of data comes in huge amount, includes various road situations, records a large number of vehicles and is easy to retrieve.

To make the best use of camera data of street traffic, we have built a 3D simulation of traffic. The simulation is conducted in Unity, a game engine that highly resembles real-world physical laws.\footnote{https://unity3d.com/} One major function of the simulator is to take in raw position information of moving vehicles and visualize them in 3D space. Another important application is to provide a reinforcement learning platform because it can take in user input to steer the vehicle and feed the results back to the deep learning program.

\section{Related Work}
Autonomous vehicles have received much attention in recent years. For heuristic approaches, Dresner proposes a reservation-based intersection control mechanism which requires a central controller that collects requests from all vehicles and returns signals of whether they can proceed or not \cite{dresner2004multiagent}. There are also solutions based on vehicle-to-vehicle and vehicle-to-infrastructure communication devices 
\cite{lee2012development} \cite{zohdy2012intersection}. However, heuristic approaches requires lots of effort to engineer the rules and make complicated algorithm. As an approximator, neural networks is able to take in images of traffic situations and output decisions. Without the need to enumerate all the rules, training the neural networks are easy and direct. Deep CNNs are now deployed in self-driving cars on highways \cite{bojarski2016end}.

However, few research work has incorporated 3D simulation or visualization. NVIDIA's simulation is based on the 2D image in front of the vehicle. Demonstration of Dresner's multiagent intersection management is by a 2D java program which simplifies the road structure and vehicles' physical properties, and does not enable steering manipulation of the vehicles \cite{dresner2004multiagent}. Krajzewicz et al.'s Simulation of Urban MObility (SUMO) is also a 2D simulation \cite{SUMO2012}. There are several constraints of these simulations. First, they are 2D views which do not observe physical properties. They can be adequate for visualization but not handy for potential reinforcement learning. Second, many of them are views from local cameras on the vehicles \cite{bojarski2016end}. They are constrained from a local perspective of the traffic situation, without auxiliary knowledge of the whole picture. 

One comparable work is the CARLA simulator released in December 2017 \cite{Dosovitskiy17}, a few months before our work. CARLA is a simulation platform that supports flexible specification of sensor suites and environmental conditions. We regard our work as a compact 3D traffic simulator between SUMO and CARLA.

Our simulation in Unity is a verisimilitude of 3D real-world situations. In the back end, raw vehicle information is generated from street camera data. Basic information includes position and direction of each vehicle, and shape and width of the roads. This can also be substituted by pre-designed data through certain algorithms. In the front end, Unity is able to turn these information into 3D simulation, and for future needs, to manipulate vehicles and detect collisions. Although the real time traffic data is processed by Python in our implementation. This simulation can be smoothly piped with any programming language that generates traffic data in the same format.

\section{Simulation in 3D Space}
\subsection{Data Structure}
Data structure of the simulation is essential for various learning and visualization purposes. The most important concepts in the simulation are road segments $S$, routes $R$, points $P$, the world map $W$ and vehicles $V$.

\paragraph{Road Segment} 
A segment $S$ is the simplest element of a route. 

Each segment has the following attributes: position, previous position, next position, scale (the width of the lane, the length of the segment, and the thickness of the road), rotation, speed limit, route id, and car id (the id of the car on that segment, if any). The position, previous position and next position are important information to locate the segment and link adjacent segments. 

There are two ways to reference each segment. One way is to look up the position in a certain route $R_i$, $i$ is the route id ($O(1)$ complexity). The other is to look the position up in the world map $W$, and then specify the route id $i$ if it is a shared segment among several routes.

\begin{figure}[h]
    \centering
    \includegraphics[width=0.5\textwidth]{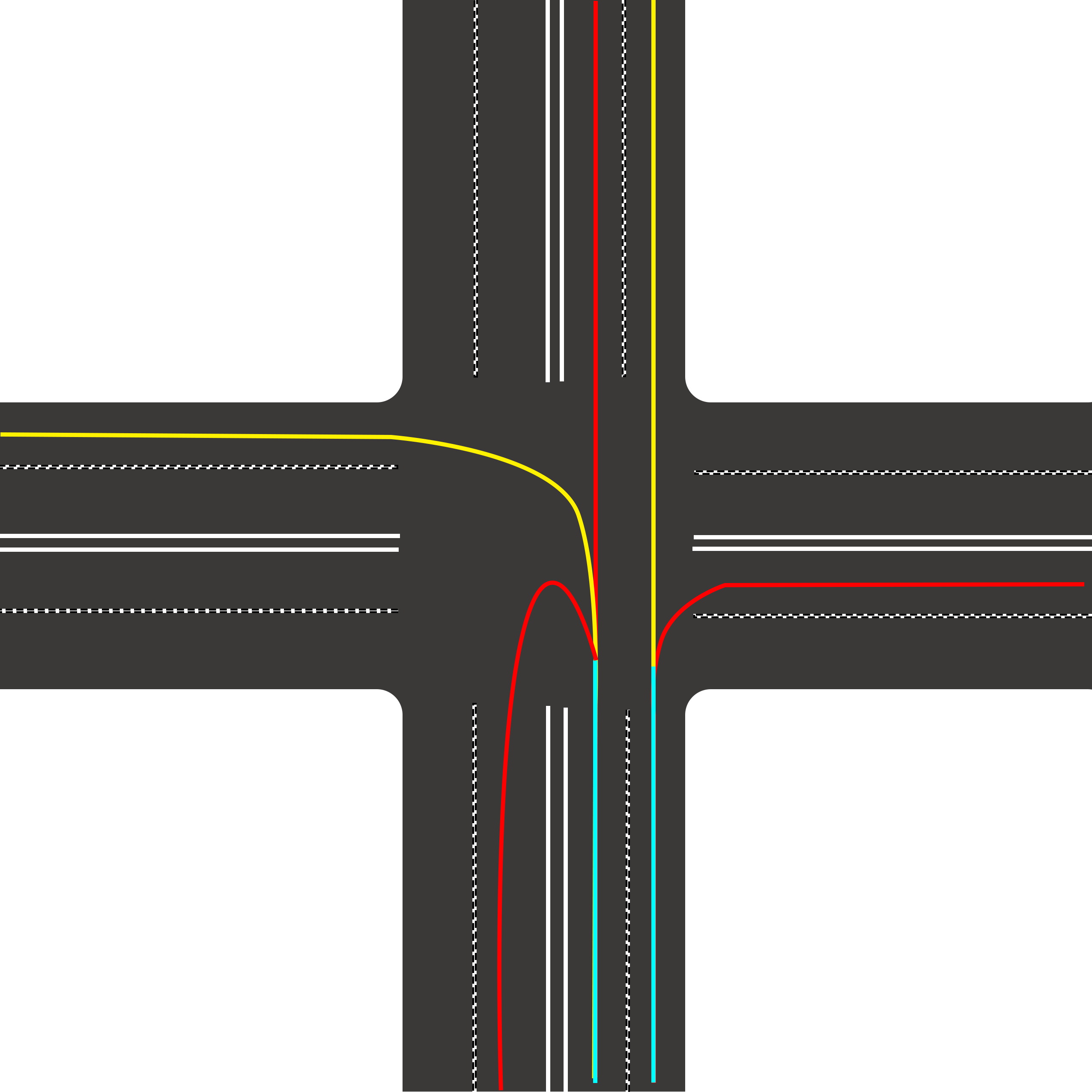}
    \caption{Routes represented in distinctive colors}
    \label{fig:inters1}
\end{figure}

\paragraph{Routes}
A route $R$ is the longest continuous line that is not overlapped. It can be either a lane of the road or a trajectory at the intersection. Figure \ref{fig:inters1} illustrates the different routes in an intersection. They should not overlap except for several shared vertices.

$R$ is a dictionary of $S$. The key is the position and the value is $S$. The set of routes is a list of $R$ indexed by route id $i$. This data structure is shown in Figure \ref{fig:datastr}

\begin{figure}[h]
    \centering
    \includegraphics[width=0.3\textwidth]{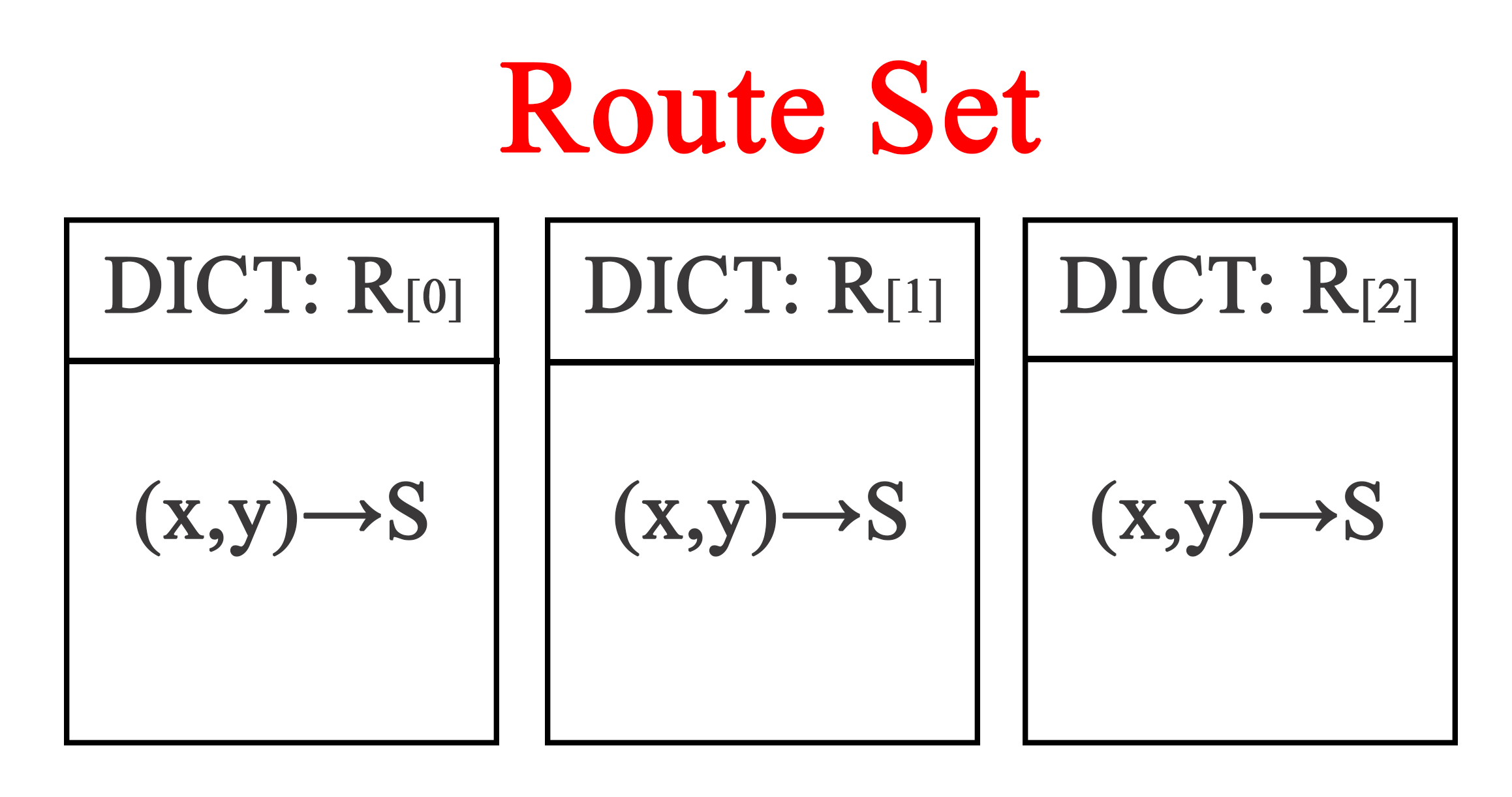}
    \caption{Relations between the set of routes $\{R_{[i]}\}$ and road segments $S$}
    \label{fig:datastr}
\end{figure}

\paragraph{Points}
Each point $P$ in the world is consists of position, list of route ids $route\_ls$ at the point, and the id of the car at the position, if any. With the position and route id, $P$ can point to a unique $S$.

The number of points is determined by how accurate the simulator is. It can also be considered as the \textit{resolution}.

An important point to note is that $route\_ls$ is ordered. The more prior a route is, any vehicle on that route has a higher priority to pass the point. The vehicles with later priority will have to wait until prior ones have passed. 

\paragraph{World Map}
The world map $W$ sized is the set of discrete points $P(x,y)$, $x, y$ are positions in the whole map with a certain \textit{resolution}.

\paragraph{Vehicle}
A vehicle $V$ has attributes like position, next position, route id, speed, and whether it is active or not. The vehicle is deactivated when it reaches the end of its route with no route change.

\subsection{Interface}
The visualization is done in the game engine Unity. The basic elements are vehicles, roads and camera.

\paragraph{Vehicles}
Vehicles, including cars, buses and police cars, are pre-designed game objects. They are assigned C\# scripts which read in the real-time position information and are updated every 1/24 seconds. If used as agents that can be manipulated by reinforcement learning, these vehicles also follow physics rules of acceleration and collision. Programs can input steering commands to manipulate the vehicles. Figure \ref{fig:datastr} shows the set of vehicle models.

\begin{figure}[h]
    \centering
    \includegraphics[width=0.4\textwidth]{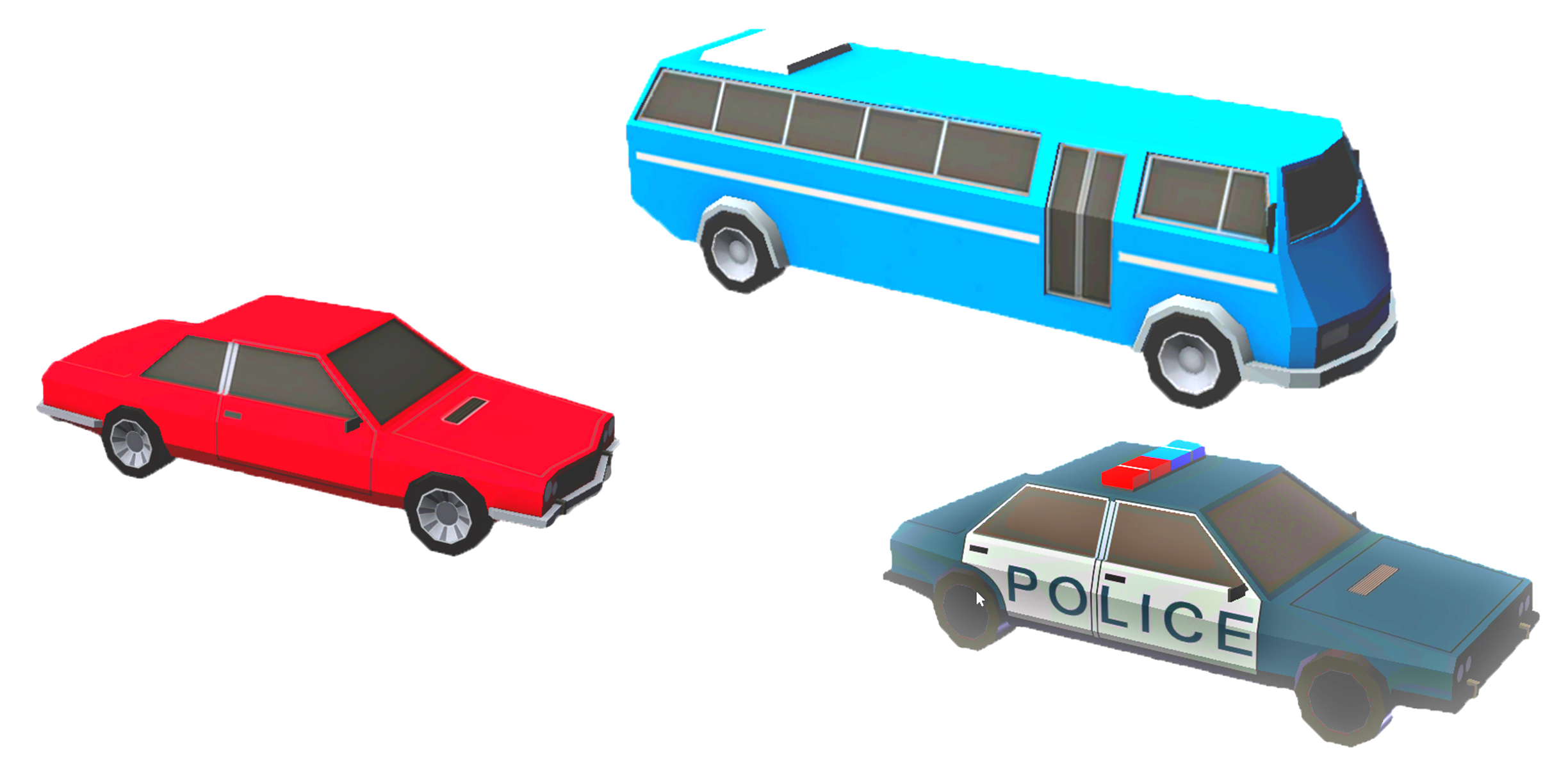}
    \caption{Models of a car, a bus and a police car in Unity}
    \label{fig:datastr}
\end{figure}

\paragraph{Roads}
Roads are formed as a group of short, linear segments. Each segment is a cube with a certain width, length and thickness. 

Roads serve as a mark of the routes, as all routes have already been stored in the inner structure. For better visual effects, there are also cubes with larger width and length which represents intersection areas.

\paragraph{Cameras}
Cameras in Unity are a powerful tool to change perspectives and highlight objects. 

For neural network training purposes, the camera can be either attached on specific vehicles or on the top of a street light to oversee the whole scene. This provides convenience for different neural network architectures. For example, the network can take in the "vision" of a vehicle and output its steering commands, and then compare that with the decision based on the whole ground truth taken from the "street camera".

Another advantage of Unity cameras is that it can render the vision by special shaders. If every game object is assigned customized shader effects, the camera can show them in different colors. It is like applying a filter on the lens of the camera. For example, the camera can assign red color to all cars, blue to all roads and so on. This greatly facilitates the CNN training, as images with different objects highlights correspond to different the input channels (such as a channel for cars, a channel for bicycles, a channel for pedestrians, etc.).

\subsection{Web Server}
The communication of Unity and real time traffic data is achieved through a web server. We set up a local Flask \footnote{Flask is a Python Microframework which can be retrieved at http://flask.pocoo.org} server at \url{localhost:5000/}. 

It has several web pages from \url{localhost:5000/car00} to \url{localhost:5000/car99} which can post the information of up to 100 cars in the traffic. This can also be scaled easily. In Unity, the user just need to initialize 100 car objects and link them to pre-written C\# scripts which takes in a variable called $url$.

It also has a web page \url{localhost:5000/line_segments}. This stores the position and rotation of basic line segments in JSON format. Unity automatically reads from this web page and initialize all roads and intersections upon starting.

\subsection{Python-Programmed Traffic Environment}

Besides the extracted data from the street cameras, we can also simulate the traffic according to certain algorithms. 

For example, the algorithm for lane following is illustrated in Algorithm \ref{alg:lanef}.

\begin{algorithm}
\caption{Lane following algorithm}\label{alg:lanef}
\begin{algorithmic}[1]

\State $\texttt{range} = \texttt{myspeed} \cdot 1$
\If {there is at least a vehicle in \texttt{range}}
\State $\texttt{dist}$ denotes the distance between my vehicle and the nearest vehicle in the front
\State $\texttt{frontspeed}$ denotes the speed of the nearest vehicle in the front
\State \If{ $\frac{\texttt{dist}}{\texttt{myspeed} - \texttt{frontspeed}} \in (0,1) $}
        \State move at \texttt{myspeed} for $t$ second
        \State move at \texttt{frontspeed} for $(1-t)$ second
        \State $\texttt{myspeed} \gets \texttt{frontspeed}$
        \EndIf
\Else
    \State move at \texttt{myspeed} for 1 second
\EndIf

\end{algorithmic}
\end{algorithm}

The algorithm for vehicles at the intersection is much more complicated. The intuition is to first check all the positions that will be passed by "my vehicle" at a certain time. If they will be occupied by vehicles with higher priority (vehicles on routes of higher priority at that time step), then "my vehicle" needs to wait. Possible challenges for decision making at an intersection are starvation, how to decide the order of vehicles to update and some other concerns.

\subsection{Examples}

\begin{figure}[h]
    \centering
    \includegraphics[width=0.5\textwidth]{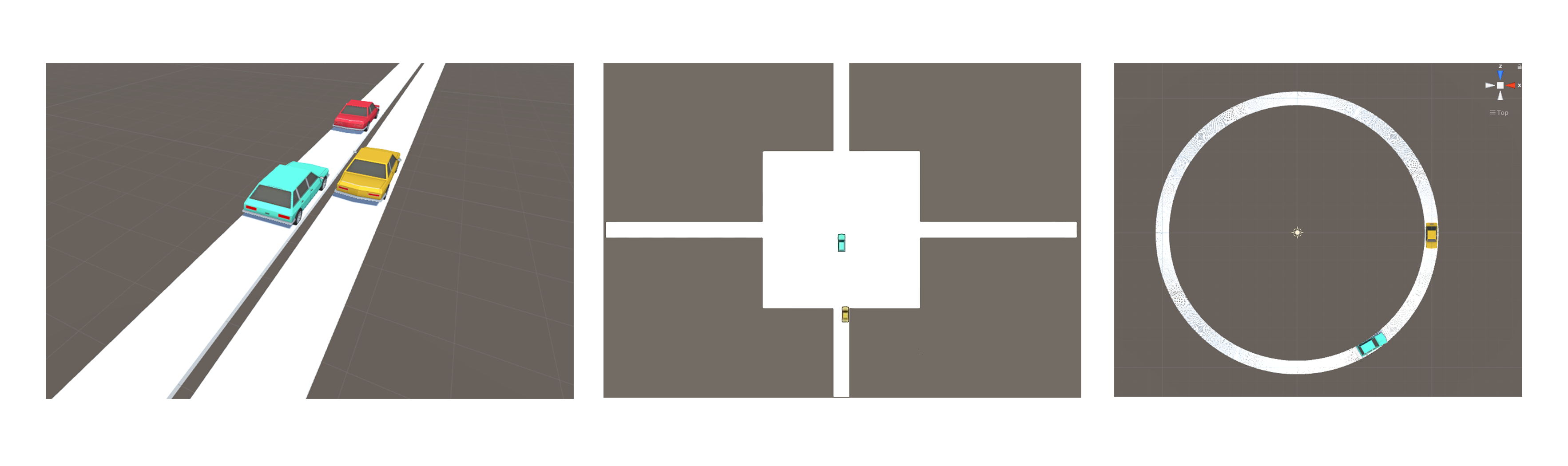}
    \caption{Samples of simulations. Left: lane following. Center: four-way intersection. Right: circular path following.}
    \label{fig:traff4}
\end{figure}

Here we demonstrate some typical examples of 3D traffic in Unity. Figure \ref{fig:traff4} shows the simulation of lane following, four-way intersection, and circular path following.

\section{Conclusion}
\paragraph{Advantages}
Compared to existent 2D simulation platforms, there are several obvious advantages of this 3D simulation. First, compared to the prevalent approach of supervised learning based on large data sets, this simulation can self-generate various traffic situations at low cost. For important edge cases, there is no longer by-chance data collection process. Variations of edge cases can be easily designed and fed into neural networks to strengthen the models. 

Second, lots of data are collected through cameras attached to vehicles, therefore only showing local views. With this system, we can easily take in data from street cameras. Street cameras can capture all agents and are almost everywhere. This is an economical and time-efficient way of data collection.

Third, this 3D simulation has wider applications than previous 2D simulations like in \cite{dresner2004multiagent} \cite{bojarski2016end}. Unlike simple programs for visualization, the simulation in Unity incorporates real world physical properties. This creates natural effects that can be useful for deep reinforcement learning.

\paragraph{Future Improvements}
This simulation still needs more improvement in the future. For example, an important feature of the data structure is that it discretizes the whole map. Every position is represented by $x$ and $y$ coordinates, so this provides inconvenience when checking whether two positions are exactly the same. Another improvement that needs to be worked on is the algorithm for vehicles at the intersection. It is hard to hand-code every rule and solve starvation and other issues at intersections. Besides, for the moment, we assume that the cars can change speed instantly. However, in the real world, it takes time to accelerate. More work needs to be done to approximate how vehicles accelerate and adjust some parts of the algorithm accordingly. More work is needed for variation of weather and road conditions.

\paragraph{Acknowledgements}
I hereby sincerely thank Camera Culture Group at MIT Media Lab which inspired me this project. Professor Ramesh Raskar and Dr. Pratik Shah supervised my work. Moreover, I appreciate the many insightful discussions with Tristan Swedish and Guy Satat. Supports from many colleagues at the lab were also important for this work.

{\small
\bibliographystyle{ieee}
\bibliography{egbib}
}

\end{document}